\documentclass[letterpaper, 10 pt, conference]{ieeeconf}  

\usepackage{float}
\usepackage{cite}
\usepackage{hyperref}
\usepackage{amsmath}
\usepackage{amssymb}
\usepackage{tikz}
\usepackage{subcaption}
\usetikzlibrary{graphs, positioning} 
\usetikzlibrary{patterns, patterns.meta}
\usetikzlibrary{calc}
\usepackage{pgfmath}
\usepackage{booktabs}
\usepackage{tabularx}

\IEEEoverridecommandlockouts                              

\title{\LARGE \bf
Spatially Intelligent Patrol Routes for \\Concealed Emitter Localization by Robot Swarms}

\author{Adam Morris$^{1}$, Timothy Pelham$^{2}$ and Edmund R. Hunt$^{1}$
\thanks{$^{1}$Adam Morris and Edmund Hunt are with School of Engineering Mathematics \& Technology, University of Bristol, UK
        {\tt\small \{adam.morris,edmund.hunt\}@bristol.ac.uk}}%
\thanks{$^{2}$Timothy Pelham is with the School of Electrical, Electronic and Mechanical Engineering, University of Bristol, UK
        {\tt\small T.G.Pelham@bristol.ac.uk}}%
}

\begin{document}

\maketitle
\thispagestyle{empty}
\pagestyle{empty}
\begin{abstract}
This paper introduces a method for designing spatially intelligent robot swarm behaviors to localize concealed radio emitters. We use differential evolution to generate geometric patrol routes that localize unknown signals independently of emitter parameters, a key challenge in electromagnetic surveillance. Patrol shape and antenna type are shown to influence information gain, which in turn determines the effective triangulation coverage. We simulate a four-robot swarm across eight configurations, assigning pre-generated patrol routes based on a specified patrol shape and sensing capability (antenna type: omnidirectional or directional). An emitter is placed within the map for each trial, with randomized position, transmission power and frequency. Results show that omnidirectional localization success rates are driven primarily by source location rather than signal properties, with failures occurring most often when sources are placed in peripheral areas of the map. Directional antennas are able to overcome this limitation due to their higher gain and directivity, with an average detection success rate of 98.75\% compared to 80.25\% for omnidirectional. Average localization errors range from 1.01--1.30 m for directional sensing and 1.67--1.90 m for omnidirectional sensing; while directional sensing also benefits from shorter patrol edges. These results demonstrate that a swarm's ability to predict electromagnetic phenomena is directly dependent on its physical interaction with the environment. Consequently, spatial intelligence, realized here through optimized patrol routes and antenna selection, is a critical design consideration for effective robotic surveillance.

\end{abstract}

\section{INTRODUCTION}
\label{sec:intro}

Concealed emitter localization (CEL) is a critical security capability that supports national security, situational awareness, and threat mitigation. For example, identifying and localizing covert radio-frequency (RF) signal sources—such as hostile communication or surveillance systems—is essential for operational effectiveness in both domestic and international theaters across government and private sector organizations. Deploying groups of mobile robotic agents to achieve distributed RF signal sensing enables rapid and accurate understanding of the electromagnetic environment, thereby enhancing technical security. Research implementations of electromagnetic sensing in multi-robot systems include, but are not limited to, signal mapping, gradient climbing, sensor coverage, surveillance, self-localization, and source localization \cite{Cortes2004CoverageNetworks, Lee2023AnSurveillance, Xu2023, Pack2009CooperativeTargets, Kitchen2020SimulatedReceivers, Sauter2020, Song2012SimultaneousRobot, Kim2014CooperativeRobots, Fink2010OnlineRobots, fraser2018radio}. 

Works addressing emitter localization often make strong assumptions about the characteristics of the target source (e.g., source power or frequency) to improve the speed, accuracy, and confidence of predictions. Concealed RF devices do not afford this luxury; thus, localization methods must operate under more relaxed assumptions. Additionally, there is a dual need: to identify the presence of unauthorized signals and to ensure that no new devices are introduced into the environment. CEL methods must therefore balance short-term signal identification with long-term maximal area coverage.  

We present a new approach to CEL that employs spatially intelligent patrolling to both accurately locate hidden emitters and provide long-term RF domain coverage. The method makes minimal assumptions about the emitter (requiring only that it be approximately isotropic) while also adapting to available sensing capabilities, whether using omnidirectional or directional antennas. We define `spatial intelligence' as a system's ability to model its physical environment, including the location and geometry of itself, its sensors, and other objects, to interpret and predict sensory data, and potentially to inform its behavioral decisions. The specific use case envisaged here is a system's internal, modeled understanding of the spatial environment being used to predict the behavior of radio signals, and using received signals to update the internal model of the environment. In the present paper behavioral decisions (i.e. patrol routes) are made offline, thus endowing the swarm with spatial intelligence before deployment.

\section{BACKGROUND}

Concealed emitter localization should make minimal assumptions about the source signal, namely agnostic to: location, transmission power and frequency, decay factor, transiency and noise level (here, we still assume an isotropic radiation pattern). As a result, it is difficult to design a robust, yet practical, search strategy that is suitable for the general case of emitters and the agent(s) (robots and possibly supporting human teammates) conducting the sensing. Many examples attempt to address such requirements using techniques such as probabilistic occupancy grids, Gaussian processes and maximum likelihood estimations \cite{Song2012SimultaneousRobot, Kim2014CooperativeRobots, Fink2010OnlineRobots, Kitchen2020SimulatedReceivers, Sauter2020}. However, each present difficulty in application to the realistic (unknown) concealed emitter localization case.

Spatio-temporal probabilistic occupancy grids (SPOG) have demonstrated effective convergence for localizing unknown, transient emitters using paired mobile robots \cite{Song2012SimultaneousRobot, Kim2014CooperativeRobots}. This approach integrates sensing models with motion planning to improve area coverage and information gain. While the method eliminates dependency on source power by using RSS ratios across robot pairs, the authors note that “$\beta$ and $\sigma^2$ [path loss and RSS measurement variance] can be obtained by calibration” \cite{Kim2014CooperativeRobots}. Thus, the method is fundamentally limited by assuming the path loss exponent, $\beta$. Deviations from the calibrated value—which can vary with signal frequency and environmental factors—lead to significant losses in localization accuracy. Such an assumption is unsuitable for CEL due to the wide variety of environments and RF device types used in covert monitoring.

Gaussian processes with maximum likelihood estimation enable online radio signal mapping and predictive modeling with uncertainty quantification \cite{Fink2010OnlineRobots}. Though effective, this technique relies on model-based priors and the associated signal source hyper-parameters (power, path loss, and even location), with performance degrading as assumptions are relaxed. Random walk-based exploration \cite{Fink2010OnlineRobots} further reduces early information gain and lacks the spatial coordination needed for multi-agent scalability.

Alternative approaches employ angle of arrival (AoA) estimation, wherein antennas sample signal time and phase differences across antenna elements to determine the most likely signal direction. Triangulation techniques can be applied thereafter; AoA estimates are aggregated across agents to compute intersection points, yielding an approximate source location. A representative example is provided by Pack et al. in the context of cooperating UAVs \cite{Pack2009CooperativeTargets}. Such techniques avoid making strict assumptions about source parameters. However, AoA is sensitive to the geometric configuration of antenna elements and requires complex signal processing algorithms (e.g. beamforming) for sufficient angular resolution. When utilizing triangulation, it is also important to consider the locations of concurrent sensor readings across multiple sensors. Pack et al. \cite{Pack2009CooperativeTargets} illustrates how similarly aligned trajectories produce degraded estimated target locations; shallow intersection angles lead to increased uncertainty over extended distances. This limitation can be partially mitigated by constraining predictions to a bounded region within the agents' collective sensing range. AoA is a powerful technique that can be more effectively exploited when spatial intelligence is properly considered, at the cost of requiring antenna arrays and complex digital signal processing techniques.

In the context of concealed emitters, we assume an isotropic radiation pattern. Antenna aperture—and thus directionality—depends on the physical size and geometry of the antenna, with focused beams often requiring large, multi-element arrays. Concealable, portable devices struggle to support such configurations due to size and orientation constraints, and typically radiate in a quasi-isotropic manner with only minor directionality. Therefore, we consider the isotropic assumption reasonable for the CEL domain, at least in this initial study.  

We employ waypoint patrolling combined with a novel triangulation formulation to monitor the RF domain and localize concealed emitters. Patrolling provides long-term behavior that is independent of source-searching actions, which is important as the presence of an emitter is not guaranteed. The patrol routes are co-optimized for both maximal area coverage with the given sensor capability and the observation geometry required for effective passive localization (c.f. Equation~(\ref{OBJ FUN}), later); the robot determines the emitter's location by analyzing the ambient signal strength of a one-way transmitter. Triangulation is preferred over gradient-following or proximity-based methods (e.g., \cite{fraser2018radio, Sauter2020}) due to constraints imposed by the physical environment and the robots themselves. The robots are not intended to remove these devices and are only required to communicate localization predictions. Furthermore, concealed devices may be physically unreachable, making methods that rely on terminating conditions based on signal strength or proximity prone to failure. The method is applied to cases where the robots are equipped either with omnidirectional or directional antennas (where a robot's orientation towards the source affects the signal fidelity).

\section{METHODOLOGY}

\subsection{Problem Statement}

A swarm of robots is tasked to monitor the RF domain of a static environment containing a concealed emitter. Robots are able to isolate the received signal strength (RSS) of the signal within some sensing range $S_R$, using either omnidirectional or directional antennas. Each robot follows a series of waypoints until all robots have completed their respective patrol routes. On completion, the sensor data is analyzed individually and/or across robots in an attempt to localize the emitter source.

\subsection{RF Model}

\subsubsection{Contemporary Equation}

The chosen sensing strategy in this investigation targets RSS, measured in decibel-milliwatts (dBm). As such, a propagation model is required. Received power from a given transmitter to a receiver can be modeled using the contemporary Friis transmission equation \cite{friis1946note}. The formula is described in Equation (\ref{FRIIS EQUATION}), both in the linear (\ref{eq1:friisa}) and logarithmic (\ref{eq1:friisb}) form, with associated units given in square brackets.

\begin{subequations}
\begin{align}
        P_{r}^{[\text{W}]} &= P_{t}^{[\text{W}]}G_{t}G_{r}\left(\frac{\lambda^{[\text{m}]}}{4\pi d^{[\text{m}]}}\right)^2 \label{eq1:friisa} \\
        P_{r}^{[\text{dBm}]} &= P_t^{[\text{dBm}]} + G_{t}^{[\text{dBi}]} + G_{r}^{[\text{dBi}]} + 20 \log_{10}\left(\frac{\lambda^{[\text{m}]}}{4\pi d^{[\text{m}]}}\right) \label{eq1:friisb}
\end{align}
\label{FRIIS EQUATION}
\end{subequations}

$P_r$ is the received power, $P_t$ is the transmitted power, $G_t$ and $G_r$ are the gain of the transmitter and receiver respectively, $\lambda$ is the wavelength of the signal ($\lambda=c/f$) and $d$ is the distance between the emitter and receiver. 

\subsubsection{Anisotropic Sensing}

Modeling the antenna gain at a specified angle requires a radiation pattern. For directional sensing, the normalized $\operatorname{sinc}$ function is used, where input $\theta$ represents the angle between the receiver line of sight and the transmitter. This is an analytically simple model that captures main and side lobe structures with beamwidth control. Converting from the linear scale to dBi requires strict positivity, therefore the $\operatorname{sinc}$ function is squared. Finally, the radiation pattern is scaled by the gain factor; the gain factor is the linear multiple by which the signal strength is multiplied at the focal point of the main lobe compared to an isotropic antenna. This term serves to both amplify the signal at the zero as well as narrow the beamwidth with increasing directionality.

For an antenna with a linear gain factor of $G_r$, the transmitted power equation is modified to include the angle-dependent parameter $G_r(\theta)$, as described in Equation~(\ref{RADIATION}). An omnidirectional antenna would have a linear gain factor of 1.0, equating to a gain of 0 dBi. 

\begin{equation}
    G_r(\theta)^{[\text{dBi}]} =
\begin{cases}
10\log_{10}\left(G_r \operatorname{sinc}^2(G_r\theta)\right) \quad G_r > 1.0 \\
0.0 \qquad \qquad G_r =1.0 \\
\mathrm{undefined} \quad G_r < 1.0
\end{cases}
    \label{RADIATION}
\end{equation} 

\subsubsection{Noise}
There are numerous methods for modeling noise generated by emitters and signal propagation through a given environment. A common approach is to combine statistical models that approximate different effects, such as fading, shadowing, and scattering, as demonstrated in \cite{Kitchen2020SimulatedReceivers}. We leave this complexity for future work, where ray-tracing-based propagation models with high-fidelity, physics-based simulations will be used for accurate multi-path and aperture modeling \cite{Pelham2023LyceanEM:Modelling}. Instead, in this work, we add normally distributed noise to the contemporary model, with a mean of zero and a variance of $0.5^2$, to represent uncertainty in RSS measurements.  

\subsubsection{Transiency}
Transiency can be modeled simply as intermittent signal transmission, where the signal strength drops out for short periods. The method depends only on the spatial features of the RSS, not on when or if a signal is received. As such, transiency can be mitigated by repeating the patrol graph until the RSS edge distribution is sufficiently populated. Transiency therefore only affects the time required to locate an emitter—often a lower priority for CEL—and not the accuracy of the prediction. For this reason, it is not modeled in this investigation.  

\subsection{Spatially Intelligent Patrol Graphs}

The proposed method utilizes triangulation as a method of localization. Triangulation has the advantage of working in both omnidirectional and directional sensing cases. Traditionally, triangulation utilizes the intersection of multiple sensors' (robots') measurements to estimate the emitter location. However, by carefully constructing the patrol route waypoints, it is possible for a single robot to generate regions of spatially explicit observation coverage.

Patrol route generation requires two robot attributes: 2-dimensional movement and a specified sensing range $S_{R}$. $S_{R}$ is an arbitrary, user-defined practical signal detection threshold. This parameter presents a range-accuracy trade-off, tuned as desired. Its definition is the distance in which a robot's sensing strategy can reasonably discern a change in received signal strength with respect to the baseline state. Setting this parameter appropriately is the key to formulating effective geometric patrol routes. 

\begin{figure}
\centering
\begin{subfigure}[b]{0.3\textwidth}
    \centering
    \begin{tikzpicture}[scale=1.5]
    \draw[fill=black, thick] (0,0.7) circle (2pt);
    \node[anchor=west] at (0.1,0.7) {Waypoint};

    \draw[thick, line width=2pt] (-0.1,0.3) -- (0.1,0.5);
    \node[anchor=west] at (0.1,0.4) {Path};

    \draw[fill=white, thick] (1.4,0.7) circle (2pt);
    \node[anchor=west] at (1.5,0.7) {Intersection};

    \draw[thick, orange, line width=2pt] (1.1,0.3) -- (1.3,0.5);
    \draw[thick, red, line width=2pt] (1.3,0.3) -- (1.5,0.5);
    \node[anchor=west] at (1.5,0.4) {Triangulation Lines};
    
\end{tikzpicture}
\end{subfigure}

\begin{subfigure}[b]{0.2\textwidth}
    \centering
    \begin{tikzpicture}[scale=1.75] 
\coordinate (A) at (-0.5,-0.289);
\coordinate (B) at (0,0.577);
\coordinate (C) at (0.5, -0.289);

\coordinate (a1) at (-1,0);
\coordinate (a2) at (-0.5,0.289);
\coordinate (a3) at (-0.5,0.866);

\coordinate (b1) at (0.5,0.866);
\coordinate (b2) at (0.5,0.289);
\coordinate (b3) at (1,0);

\coordinate (c1) at (0.5,-0.866);
\coordinate (c2) at (0,-0.577);
\coordinate (c3) at (-0.5,-0.866);


\draw[draw=none, pattern color=orange, pattern={Lines[angle=-30,distance=3pt,line width=1pt]}, opacity=0.75, line width=1pt]
    (a1) -- (c1) -- (b3) -- (a3) -- cycle;
    
\draw[draw=none, pattern color=orange, pattern={Lines[angle=90,distance=3pt,line width=1pt]}, opacity=0.75, line width=1pt]
    (c3) -- (a3) -- (b1) -- (c1) -- cycle;

\draw[draw=none, pattern color=orange, pattern={Lines[angle=30,distance=3pt,line width=1pt]}, opacity=0.75, line width=1pt]
    (a1) -- (b1) -- (b3) -- (c3) -- cycle;

\draw[thick, line width=2pt]
    (A) -- (B) -- (C) -- cycle;

\draw[thick]
    (A) -- (a1) -- (a2) -- (a3) -- (B);

\draw[thick]
    (B) -- (b1) -- (b2) -- (b3) -- (C);

\draw[thick]
    (C) -- (c1) -- (c2) -- (c3) -- (A);

\foreach \point in {A, B, C}
    \fill[black] (\point) circle (2pt);

\foreach \point in {A, a1, a2, a3, B, b1, b2, b3, C, c1, c2, c3}
    \draw[fill=white, thick] (\point) circle (1pt); 
    
\end{tikzpicture}
    \caption{Min $S_R=\frac{2 \sqrt{3}}{3}l$ meters.}
    \label{UNAMED 1}
\end{subfigure}
\hspace{0.5cm}
\vspace{0.4cm}
\begin{subfigure}[b]{0.2\textwidth}
    \centering
    \begin{tikzpicture}[scale=0.75] 
\coordinate (A) at (-0.866,-0.5);
\coordinate (B) at (0,1);
\coordinate (C) at (0.866, -0.5);

\coordinate (a1) at (-0.867,1.5);
\coordinate (a2) at (0.0,3.0);
\coordinate (a3) at (0.867, 0.5);
\coordinate (a4) at (1.732, 2.0);

\coordinate (b1) at (0.0, -1.0);
\coordinate (b2) at (0.867,-2.5);
\coordinate (b3) at (1.732, 0.0);
\coordinate (b4) at (2.598, -1.5);

\coordinate (c1) at (-0.866,0.5);
\coordinate (c2) at (-2.598,0.5);
\coordinate (c3) at (-0.866,-1.5);
\coordinate (c4) at (-2.598,-1.5);

\draw[black, pattern color=red, pattern={Lines[angle=90,distance=3pt,line width=1pt]}, opacity=1, line width=1pt]
    (A) -- (a1) -- (a2) -- (B) -- cycle;
    
\draw[black, pattern color=orange, pattern={Lines[angle=30,distance=3pt,line width=1pt]}, opacity=1, line width=1pt]
    (A) -- (a3) -- (a4) -- (B) -- cycle;

\draw[black, pattern color=orange, pattern={Lines[angle=90,distance=3pt,line width=1pt]}, opacity=1, line width=1pt]
    (B) -- (b1) -- (b2) -- (C) -- cycle;

\draw[black, pattern color=red, pattern={Lines[angle=-30,distance=3pt,line width=1pt]}, opacity=1, line width=1pt]
    (B) -- (b3) -- (b4) -- (C) -- cycle;

\draw[black, pattern color=orange, pattern={Lines[angle=-30,distance=3pt,line width=1pt]}, opacity=1, line width=1pt]
    (C) -- (c1) -- (c2) -- (A) -- cycle;
    
\draw[black, pattern color=red, pattern={Lines[angle=30,distance=3pt,line width=1pt]}, opacity=1, line width=1pt]
    (C) -- (c3) -- (c4) -- (A) -- cycle;



\draw[thick, line width=2pt]
    (A) -- (B) -- (C) -- cycle;

\foreach \point in {A, B, C}
    \fill[black] (\point) circle (3pt);

\end{tikzpicture}
    \caption{Min $S_R = \frac{2}{\sqrt3}l$ meters.}
    \label{UNAMED 5}
\end{subfigure}
\vspace{0.4cm}
\begin{subfigure}[b]{0.2\textwidth}
    \centering
    \begin{tikzpicture}[scale=1.75] 
\coordinate (A) at (-0.5,-0.5);
\coordinate (B) at (-0.5,0.5);
\coordinate (C) at (0.5, 0.5);
\coordinate (D) at (0.5, -0.5);

\coordinate (a) at (-1,-0.5);
\coordinate (b) at (-1,0.5);
\coordinate (c) at (1, 0.5);
\coordinate (d) at (1, -0.5);

\coordinate (e) at (-0.5,-1);
\coordinate (f) at (-0.5,1);
\coordinate (g) at (0.5, 1);
\coordinate (h) at (0.5, -1);


\draw[draw=none, pattern color=orange, pattern={Lines[angle=0,distance=3pt,line width=1pt]}, opacity=1, line width=1pt]
    (a) -- (b) -- (c) -- (d) -- cycle;
    
\draw[draw=none, pattern color=orange, pattern={Lines[angle=90,distance=3pt,line width=1pt]}, opacity=1, line width=1pt]
    (e) -- (f) -- (g) -- (h) -- cycle;

\draw[draw=none]
    (A) -- (B) -- (C) -- (D) -- cycle;

\draw[thick, line width=2pt]
    (A) -- (B) -- (C) -- (D) -- cycle;

\foreach \point in {A, B, C, D}
    \fill[black] (\point) circle (2pt);

\foreach \point in {A, B, C, D}
    \draw[fill=white, thick] (\point) circle (1pt); 
    
\end{tikzpicture}
    \caption{Min $S_R=\frac{d}{2}$ meters.}
    \label{UNAMED 2}
\end{subfigure}
\hspace{0.5cm}
\begin{subfigure}[b]{0.2\textwidth}
    \centering
    \begin{tikzpicture}[scale=0.75] 
\coordinate (A) at (-0.707,-0.707);
\coordinate (B) at (-0.707,0.707);
\coordinate (C) at (0.707, 0.707);
\coordinate (D) at (0.707, -0.707);

\coordinate (a1) at (-2.121,0.707);
\coordinate (a2) at (-2.121,2.121);
\coordinate (a3) at (0.707, 0.707);
\coordinate (a4) at (0.707, 2.121);

\coordinate (b1) at (0.707, 2.121);
\coordinate (b2) at (2.121,2.121);
\coordinate (b3) at (0.707, -0.707);
\coordinate (b4) at (2.121, -0.707);

\coordinate (c1) at (-0.707, -0.707);
\coordinate (c2) at (-0.707, -2.121);
\coordinate (c3) at (2.121, -0.707);
\coordinate (c4) at (2.121, -2.121);

\coordinate (d1) at (-0.707, 0.707);
\coordinate (d2) at (-2.121, 0.707);
\coordinate (d3) at (-0.707, -2.121);
\coordinate (d4) at (-2.121, -2.121);

\draw[black, pattern color=orange, pattern={Lines[angle=-45,distance=3pt,line width=1pt]}, opacity=1, line width=1pt]
    (A) -- (a1) -- (a2) -- (B) -- cycle;

\draw[black, pattern color=orange, pattern={Lines[angle=45,distance=3pt,line width=1pt]}, opacity=1, line width=1pt]
    (B) -- (b1) -- (b2) -- (C) -- cycle;

\draw[black, pattern color=orange, pattern={Lines[angle=-45,distance=3pt,line width=1pt]}, opacity=1, line width=1pt]
    (C) -- (c3) -- (c4) -- (D) -- cycle;

\draw[black, pattern color=orange, pattern={Lines[angle=45,distance=3pt,line width=1pt]}, opacity=1, line width=1pt]
    (D) -- (d3) -- (d4) -- (A) -- cycle;

\draw[black, pattern color=red, pattern={Lines[angle=-45,distance=3pt,line width=1pt]}, opacity=1, line width=1pt]
    (D) -- (d1) -- (d2) -- (A) -- cycle;

\draw[black, pattern color=red, pattern={Lines[angle=45,distance=3pt,line width=1pt]}, opacity=1, line width=1pt]
    (A) -- (a3) -- (a4) -- (B) -- cycle;

\draw[black, pattern color=red, pattern={Lines[angle=45,distance=3pt,line width=1pt]}, opacity=1, line width=1pt]
    (C) -- (c1) -- (c2) -- (D) -- cycle;

\draw[black, pattern color=red, pattern={Lines[angle=-45,distance=3pt,line width=1pt]}, opacity=1, line width=1pt]
    (B) -- (b3) -- (b4) -- (C) -- cycle;
    
\draw[thick, line width=2pt]
    (A) -- (B) -- (C) -- (D) -- cycle;

\foreach \point in {A, B, C, D}
    \fill[black] (\point) circle (3pt);

    
\end{tikzpicture}
    \caption{Min $S_R = \sqrt2 d$ meters.}
    \label{UNAMED 6}
\end{subfigure}
\vspace{0.4cm}
\begin{subfigure}[b]{0.2\textwidth}
    \centering
    \begin{tikzpicture}[scale=1.25] 
\coordinate (A) at (-0.5,-0.866);
\coordinate (B) at (-1,0);
\coordinate (C) at (-0.5, 0.866);
\coordinate (D) at (0.5, 0.866);
\coordinate (E) at (1, 0);
\coordinate (F) at (0.5, -0.866);

\coordinate (A1) at (-0.5,-1.366);
\coordinate (A2) at (-0.933,-1.116);
\coordinate (B1) at (-1.433,-0.25);
\coordinate (B2) at (-1.433,0.25);
\coordinate (C1) at (-0.933, 1.116);
\coordinate (C2) at (-0.5, 1.366);
\coordinate (D1) at (0.5, 1.366);
\coordinate (D2) at (0.933, 1.116);
\coordinate (E1) at (1.433, 0.25);
\coordinate (E2) at (1.433, -0.25);
\coordinate (F1) at (0.933, -1.116);
\coordinate (F2) at (0.5, -1.366);

\coordinate (a1) at (-0.5, -0.289);
\coordinate (b1) at (-0.5, 0.289);
\coordinate (c1) at (0, 0.577);
\coordinate (d1) at (0.5, 0.289);
\coordinate (e1) at (0.5, -0.289);
\coordinate (f1) at (0, -0.577);

\draw[white, line width=0pt, pattern color=orange, pattern={Lines[angle=30,distance=3pt,line width=1pt]}, opacity=0.75]
    (A2) -- (B1) -- (D2) -- (E1) -- cycle;

\draw[white, line width=0pt, pattern color=orange, pattern={Lines[angle=-30,distance=3pt,line width=1pt]}, opacity=0.75]
    (B2) -- (C1) -- (E2) -- (F1) -- cycle;

\draw[white, line width=0pt, pattern color=orange, pattern={Lines[angle=90,distance=3pt,line width=1pt]}, opacity=0.75]
    (C2) -- (D1) -- (F2) -- (A1) -- cycle;

Draw the polygon
\draw[thick, line width=2pt]
    (A) -- (B) -- (C) -- (D) -- (E) -- (F) --  cycle;

\draw[thick]
    (A) -- (a1) -- (B) -- (b1) -- (C) --(c1) -- (D) -- (d1) -- (E) -- (e1) -- (F) -- (f1) --  cycle;

\foreach \point in {A, B, C, D, E, F}
    \fill[black] (\point) circle (2.5pt);

\foreach \point in {A, a1, B, b1, C, c1, D, d1, E, e1, F, f1}
    \draw[fill=white, thick] (\point) circle (1.5pt); 
    
\end{tikzpicture}
    \caption{Min $S_R=\frac{\sqrt3}{2}s$ meters.}
    \label{UNAMED 3}
\end{subfigure}
\hspace{0.5cm}
\begin{subfigure}[b]{0.2\textwidth}
    \centering
    \begin{tikzpicture}[scale=0.65] 
\coordinate (A) at (-0.5,-0.866);
\coordinate (B) at (-1,0);
\coordinate (C) at (-0.5, 0.866);
\coordinate (D) at (0.5, 0.866);
\coordinate (E) at (1, 0);
\coordinate (F) at (0.5, -0.866);

\coordinate (a1) at (-2.500, -0.866);
\coordinate (a2) at (-3.000, 0.000);
\coordinate (a3) at (0.500, 0.866);
\coordinate (a4) at (0.000, 1.732);

\coordinate (b1) at (1.000, 0.000);
\coordinate (b2) at (1.500, 0.866);
\coordinate (b3) at (-2.000, 1.732);
\coordinate (b4) at (-1.500, 2.598);

\coordinate (c1) at (0.500, 2.598);
\coordinate (c2) at (1.500, 2.598);
\coordinate (c3) at (0.500, -0.866);
\coordinate (c4) at (1.500, -0.866);

\coordinate (d1) at (2.500, 0.866);
\coordinate (d2) at (3.000, 0.000);
\coordinate (d3) at (-0.500, -0.866);
\coordinate (d4) at (0.000, -1.732);

\coordinate (e1) at (-1.000, 0.000);
\coordinate (e2) at (-1.500, -0.866);
\coordinate (e3) at (2.000, -1.732);
\coordinate (e4) at (1.500, -2.598);

\coordinate (f1) at (-0.500, 0.866);
\coordinate (f2) at (-1.500, 0.866);
\coordinate (f3) at (-0.500, -2.598);
\coordinate (f4) at (-1.500, -2.598);

\draw[black, pattern color=orange, pattern={Lines[angle=0,distance=3pt,line width=1pt]}, opacity=1, line width=1pt]
    (A) -- (a1) -- (a2) -- (B) -- cycle;

\draw[black, pattern color=orange, pattern={Lines[angle=-60,distance=3pt,line width=1pt]}, opacity=1, line width=1pt]
    (B) -- (b3) -- (b4) -- (C) -- cycle;

\draw[black, pattern color=orange, pattern={Lines[angle=60,distance=3pt,line width=1pt]}, opacity=1, line width=1pt]
    (C) -- (c1) -- (c2) -- (D) -- cycle;

\draw[black, pattern color=orange, pattern={Lines[angle=0,distance=3pt,line width=1pt]}, opacity=1, line width=1pt]
    (D) -- (d1) -- (d2) -- (E) -- cycle;

\draw[black, pattern color=orange, pattern={Lines[angle=-60,distance=3pt,line width=1pt]}, opacity=1, line width=1pt]
    (E) -- (e3) -- (e4) -- (F) -- cycle;

\draw[black, pattern color=orange, pattern={Lines[angle=60,distance=3pt,line width=1pt]}, opacity=1, line width=1pt]
    (F) -- (f3) -- (f4) -- (A) -- cycle;

\draw[black, pattern color=red, pattern={Lines[angle=60,distance=3pt,line width=1pt]}, opacity=1, line width=1pt]
    (A) -- (a3) -- (a4) -- (B) -- cycle;

\draw[black, pattern color=red, pattern={Lines[angle=-60,distance=3pt,line width=1pt]}, opacity=1, line width=1pt]
    (F) -- (f1) -- (f2) -- (A) -- cycle;

\draw[black, pattern color=red, pattern={Lines[angle=0,distance=3pt,line width=1pt]}, opacity=1, line width=1pt]
    (E) -- (e1) -- (e2) -- (F) -- cycle;

\draw[black, pattern color=red, pattern={Lines[angle=60,distance=3pt,line width=1pt]}, opacity=1, line width=1pt]
    (D) -- (d3) -- (d4) -- (E) -- cycle;

\draw[black, pattern color=red, pattern={Lines[angle=-60,distance=3pt,line width=1pt]}, opacity=1, line width=1pt]
    (C) -- (c3) -- (c4) -- (D) -- cycle;

\draw[black, pattern color=red, pattern={Lines[angle=0,distance=3pt,line width=1pt]}, opacity=1, line width=1pt]
    (B) -- (b1) -- (b2) -- (C) -- cycle;

\draw[thick, line width=2pt]
    (A) -- (B) -- (C) -- (D) -- (E) -- (F) --  cycle;

\foreach \point in {A, B, C, D, E, F}
    \fill[black] (\point) circle (3pt);

    
\end{tikzpicture}
    \caption{Min $S_R = 2s$ meters.}
    \label{UNAMED 7}
\end{subfigure}

\caption{Spatially explicit observation coverage for omnidirectional (column 1) and directional (column 2) sensing geometric patrol routes. (a) (b): Triangular (c) (d): Square (e) (f): Hexagonal. Cross-hatched areas are regions where if an emitter were present an exact source location can be extrapolated. Internal orthogonal lines drawn to their minimum required length. External coverage in reality would extend equally as far out as the minimum distance required internally. Directional lines are drawn to the minimum required length for full coverage, with slope equal to the antenna offsets $\pm \psi$. Triangulation line colors represent coverage per antenna. Rotation coverage at the vertices for directional is not shown.}
\label{OBSERVATION AREA COVERAGE}
\end{figure}
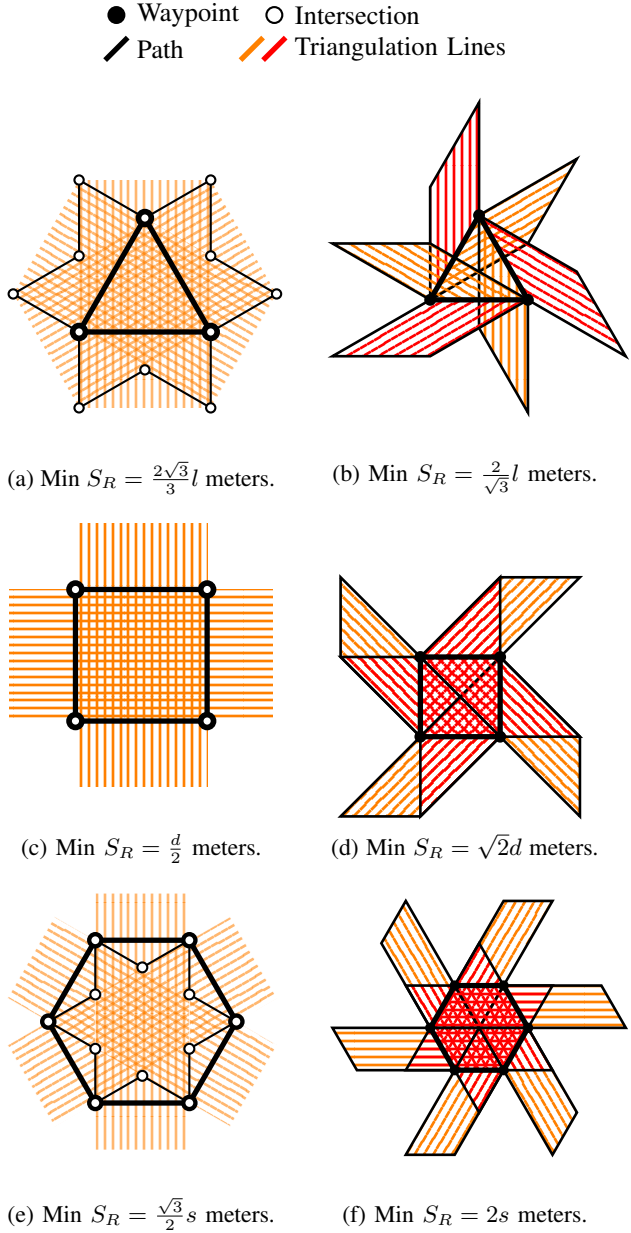

\subsubsection{Omnidirectional Sensing}

Information coverage is dependent on the sensing strategy. The first case considers omnidirectional sensing (in the azimuth plane), wherein the orientation of the robot does not affect the RSS. Due to an assumed radial radiation pattern of a concealed emitter, any straight path across the environment must have a single maximum region. The point at which this occurs must be orthogonal to the source location (except if the maximum lies on the vertices). Additional patrol edges, offset at some angle to the previous edge, create regions of triangulation based on the maximum strength indexes. Any emitter placed within these regions will produce a local maximum on each of the edge signal strength distributions, hence an exact location can be determined. The extent of this area is bounded by both the length of the edge and the sensing range. This technique hereafter is referred to as edge maxima triangulation. As we are solely exploring the change in strength across an edge, the method removes the dependency on knowing the absolute transmission power of the emitter.  

The possibility of localizing an emitter via a series of connected straight edges inspires the use of regular, geometric patrol shapes. Three routes are initially considered: triangular, square and hexagonal. It can be determined that the boundary of the triangulation region is constructed by the enclosed shape of all the intersection points of orthogonal lines (to the edge bearings) at the vertices. Fig. \ref{OBSERVATION AREA COVERAGE} (\ref{UNAMED 1}, \ref{UNAMED 2}, \ref{UNAMED 3}) illustrates the coverage for each of the respective patrol shapes. It further notes the minimum required sensing range for maximal coverage given a patrol edge length ($l$, $d$ or $s$ respectively). Beyond this length the area coverage degrades.

\subsubsection{Directional Sensing}

Edge maxima triangulation can be further utilized for directional sensing. Directional antennas produce the highest gain at orientations directly in the line of sight of the emitter. Therefore, the observation lines are shifted in line with the antenna offset $\psi$ (with respect to the robot). Furthermore, the sensing range is conservatively assumed to be the practical sensitivity of the main lobe (i.e., within the beamwidth).

Depending on the gain of the antenna, the sensing range is increased at the expense of narrowing the observation field of view with increasing directionality. As such, multiple antennas are often used in conjunction to alleviate the reduced coverage. Therefore, in the present investigation two identical directional antennas are used, offset symmetrically $\pm \psi$ degrees from the robot's front facing orientation. 

Extending edge signal strength maxima triangulation for each patrol shape to two directional antennas gives the explicit observation coverages seen in Fig. \ref{OBSERVATION AREA COVERAGE} (\ref{UNAMED 5}, \ref{UNAMED 6}, \ref{UNAMED 7}) (for a robot traveling clockwise around the patrol graph). The antenna offset $\psi$ is uniquely calculated for each route based on the angle from the vertices to the shape centroid, ensuring there are no blind spots within the enclosed patrol graph.

When utilizing directional sensing (Fig. \ref{UNAMED 5}, \ref{UNAMED 6}, \ref{UNAMED 7}), there is additional information coverage at the vertices (not illustrated). As a robot rotates towards the next waypoint, antennas cover an additional sector swept by the change in orientation. Each respective shape produces varying coverages due to the antenna offsets and angle between patrol edges. These sectors are considered in the multi-robot case, where robots can share their observations with one another to make localization predictions. Vertex RSS distributions provide an additional triangulation line based on the data collected solely at the given waypoint during its change in orientation. 

\subsection{Swarm Behavior Design}

Robots are to be distributed strategically across the environment to maximize total spatially explicit area coverage. Such effective configurations may be discovered using differential evolution (DE). DE offers a straightforward implementation of a multi-objective minimization task based on biologically inspired evolutionary adaptations. Applications of DE can be seen within swarm robotics, for example illustrating its effectiveness at improving collaborative target detection in swarms of drones \cite{cimino2016using}. Equation~(\ref{OBJ FUN}) outlines an objective function to be minimized using this approach, and its metrics are described in Table \ref{PARAMETERS}.

\begin{gather}
        \min{f}: \mathbb{R}^{(\text{$p$}, \text{$4$})} \to \mathbb{R} \notag \\
    f(\mathbf{x}) = - \text{$w_1$} \alpha + \text{$w_2$} \beta - \text{$w_3$}\gamma - \text{$w_4$}\delta + \eta
\label{OBJ FUN}
\end{gather}

\begin{table}[h]
\centering
\begin{tabular}{l p{3.3cm} p{3.3cm}}
\hline
\textbf{Metrics} & \textbf{Description} & \textbf{Objective} \\
\hline
$\alpha$ & Measures the spatially explicit coverage as a percentage of the map & Maximize the proportion of the map where exact localization is possible \\

$\beta$ & Quantifies the spatial overlap between agents' spatially explicit regions & Minimize overlap between agents to promote efficient, non-redundant coverage \\

$\gamma$ & Measures the multi-agent triangulation potential between patrol regions & Maximize overlap in regions where combined agent sensing can achieve localization not possible individually \\

$\delta$ & Evaluates the angular diversity of patrol graph rotations & Encourage varied route orientations to improve information gain and triangulation coverage \\

$\eta$ & Represents a hard penalty for invalid patrol configurations & Penalize routes that extend beyond the map boundary or whose edges intersect with it \\
\hline
\end{tabular}
\caption{Metric definitions and objectives for DE-based patrol optimization. Each metric is normalized as a score between 0 and 1 (0-100\%) calculated as a function of the candidate solution.}
\label{PARAMETERS}
\end{table}

In our DE implementation, a candidate within the population $\mathbf{x}$, of size $p$, contains 4 decision variables per robot. These include the patrol shape centroid coordinates, $(x_c, y_c)$, rotations, $\omega$, and edge lengths, $s(S_R)$. Edge length is bounded between a specified minimum (in this investigation this is set to 1 meter) and the maximum length as a function of the sensing range, the inverse of which is outlined in Fig.~\ref{OBSERVATION AREA COVERAGE} per patrol shape. The hyperparameters of the system include $w_{1-4}$, being the weighting towards each of the objective function metrics, as well as the assigned patrol shape for individual robots. 

\label{sec:swarm}

\subsection{Emitter Localization Using Least Squares}

Once a maximum strength reading is identified, a triangulation line is formed as a straight line passing through the coordinate $(x_0, y_0)$ of the associated strength reading, with slope $\phi$. For omnidirectional sensing, $\phi$ is the perpendicular gradient to the patrol edge direction (from the starting vertex to the next waypoint). In the directional case, $\phi$ corresponds to the antenna’s orientation in the global reference frame. If at least two lines can be constructed, the intersection point is estimated using a least-squares minimization, as formulated in Equation~(\ref{equ:lse}). 

\begin{equation}
\label{equ:lse}
\begin{split}
\underbrace{\begin{bmatrix}
-\sin \phi_1 & \cos \phi_1 \\
\vdots & \vdots \\
-\sin \phi_m & \cos \phi_m
\end{bmatrix}}_{\displaystyle \mathbf{A}}
\underbrace{\begin{bmatrix} x \\ y \end{bmatrix}}_{\displaystyle \mathbf{p}}
&=
\underbrace{\begin{bmatrix}
-x_{01}\sin \phi_1 + y_{01}\cos \phi_1 \\
\vdots \\
-x_{0m}\sin \phi_m + y_{0m}\cos \phi_m
\end{bmatrix}}_{\displaystyle \mathbf{b}} \\[4pt]
\min 
J(\mathbf{p}) &= \| \mathbf{A} \mathbf{p} - \mathbf{b} \|^2 \\[4pt]
\boxed{\mathbf{p} = (\mathbf{A}^\top \mathbf{A})^{-1} \mathbf{A}^\top \mathbf{b}}, 
\quad \mathbf{p} &= \begin{bmatrix} x \\ y \end{bmatrix}, \quad m \geq 2
\end{split}
\end{equation}

\subsection{Simulation Setup}

In our simulated application of these methods, a swarm of four agents is placed in a static, empty map. Four patrol route shape configurations are used to generate waypoints for both sensing strategies, resulting in eight total simulation scenarios. These configurations include three homogeneous cases (all agents using the same shape) and one mixed case (two triangular, one square, and one hexagonal). For each scenario, a source node is randomly generated and placed within the map boundary. Source parameters are drawn from a uniform distribution: position coordinates $(x, y)$ are bounded by the map dimensions, transmission power is bounded within $[20, 30]$ dBm, and frequency within $[0.4, 2.4]$ GHz. Robots complete a single traversal of their respective patrol graphs. Once all robots finish, the collected data is combined to produce a localization prediction. Each scenario is repeated for 100 trials, recording whether a trial is successful—defined as generating two or more triangulation lines whose intersection point lies within the map boundary—and calculating the absolute error (in meters) between the predicted and true source positions. 

Rearranging the propagation model in terms of distance, $d$, allows an approximate sensing range, $S_r$, to be determined based on a desired received power threshold, $P_r = P_{thresh}$. The transmit power, $P_t$, and wavelength, $\lambda = \frac{c}{f}$, can be adjusted to provide either a conservative or optimistic estimate of range for the agents. Here, we use a mid-range estimate by taking the midpoint of the uniform parameter bounds and a threshold value of $-30$dBm, however this choice is arbitrary. Antenna gain is also incorporated into this calculation, meaning that antennas with higher signal fidelity can support larger patrol routes without compromising internal area coverage. For the present simulations, directional antennas have a focal point gain of 8dBi. Finally, the differential evolution hyperparameters are manually tuned for each shape configuration. This tuning reflects the varying spatially explicit observation regions and single detection points, providing the best chance of producing centralized localization predictions. In general, the values for $w_{1-4}$ fell within the range of 1.0 to 5.0. Future work could improve patrol route generation through formalized hyperparameter optimization.

\section{RESULTS}

\subsection{Omnidirectional Versus Directional Sensing}

\begin{figure}
    \centering
    \includegraphics[width=\linewidth]{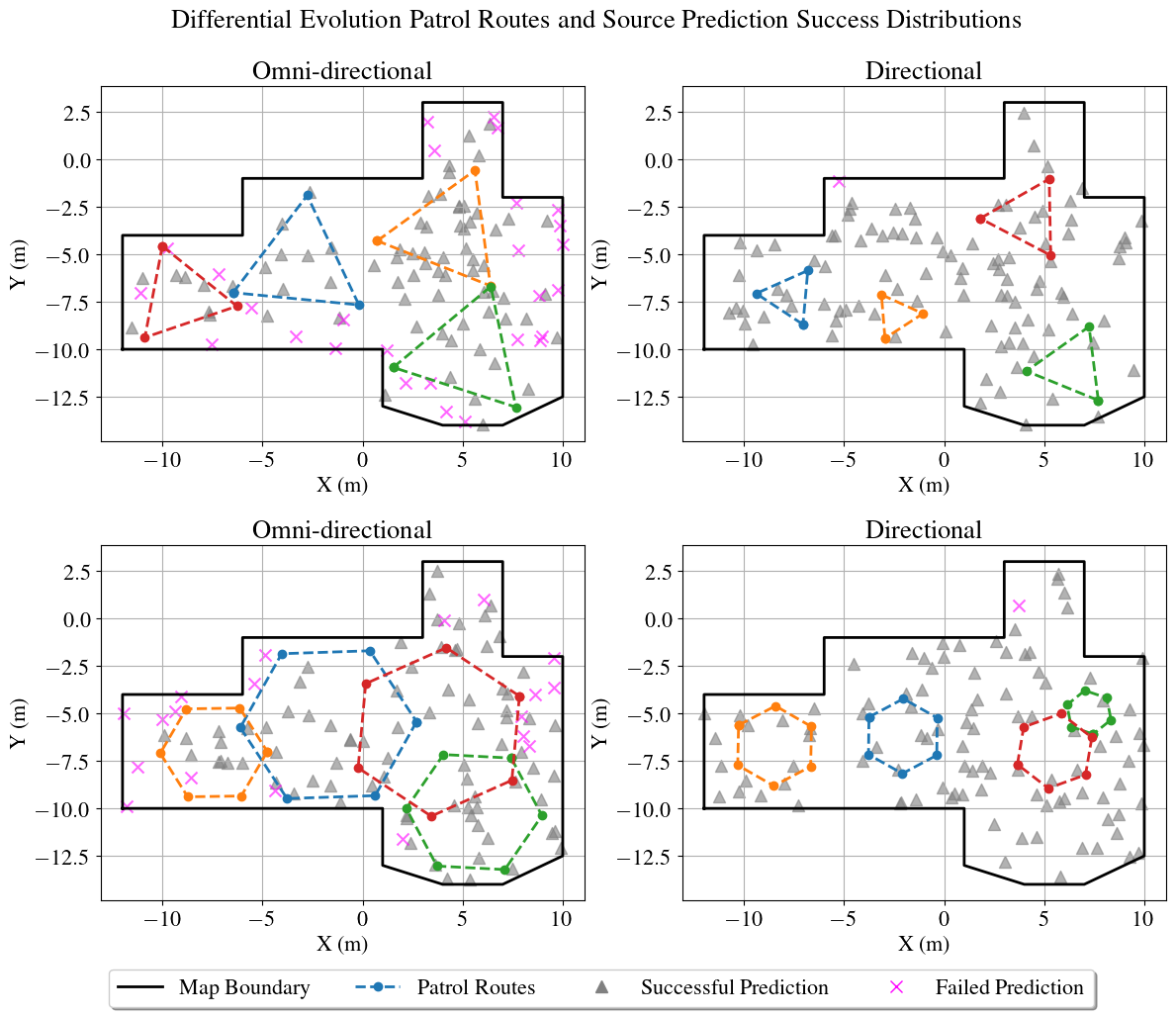}
    \caption{Waypoint generation for 4 triangular and 4 hexagonal patrol shapes utilizing each of the sensing strategies. Randomly generated source locations across the 100 trials are included, along with the whether a successful localization prediction was made by the agents.}
    \label{fig:patrols}
\end{figure}

\begin{figure}
    \centering
    \includegraphics[width=\linewidth]{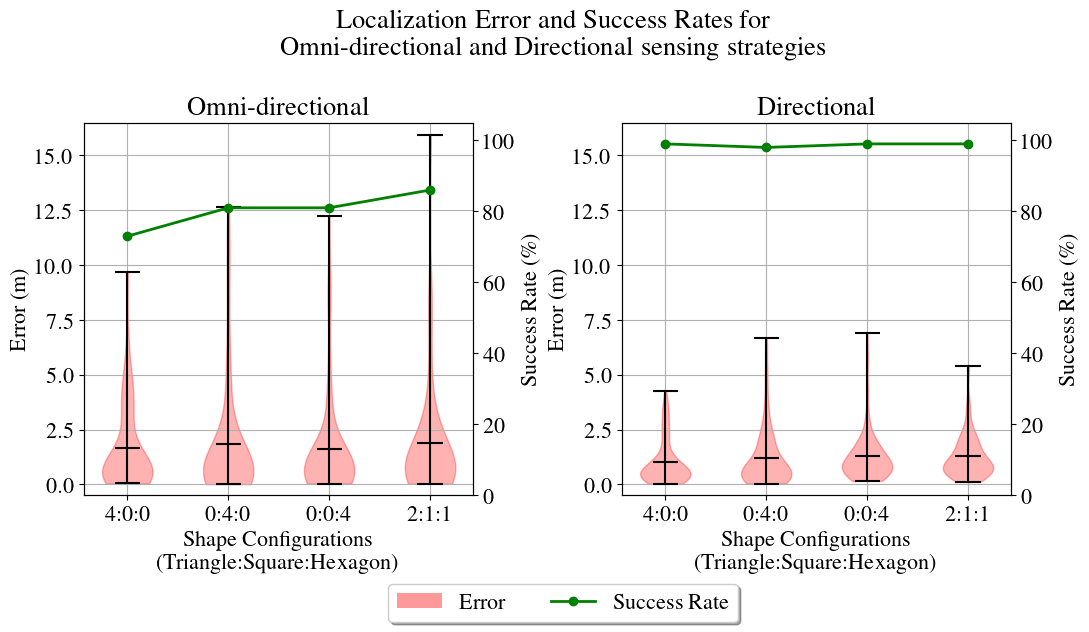}
    \caption{Localization error between predicted and true source location, and percentage success rate of forming a successful prediction for both sensing strategies across each patrol shape configuration.}
    \label{fig:error}
\end{figure}

Figure \ref{fig:patrols} shows the generated waypoint patrol routes for the triangular and hexagonal shape configurations, for the respective sensing strategies (a single omnidirectional or dual directional antennas). Directional antennas produce shorter patrol edges while maintaining sufficient area coverage. As shown in Figure \ref{fig:error}, directional sensing achieves successful localization predictions more frequently and with greater accuracy than its omnidirectional counterpart. A root cause of this discrepancy is the number of possible triangulation lines that each strategy can generate. For omnidirectional sensing, the number of triangulation lines is limited to the total number of patrol edges. In the triangular configuration, this yields a maximum of $4 \times 3 = 12$ lines. In contrast, directional sensing provides an additional RSS distribution at each vertex, producing one extra triangulation line per vertex. These vertex readings, combined with the edge readings, are then doubled by the presence of a second antenna. Consequently, for the same patrol route geometry, directional sensing offers up to $4 \times (3 + 3) \times 2 = 48$ possible line constructions. Combined with higher antenna gain, this greatly mitigates the adverse effects of noisy RSS readings, enabling more accurate and consistent localization predictions.

\begin{figure}
    \centering
    \includegraphics[width=1\linewidth]{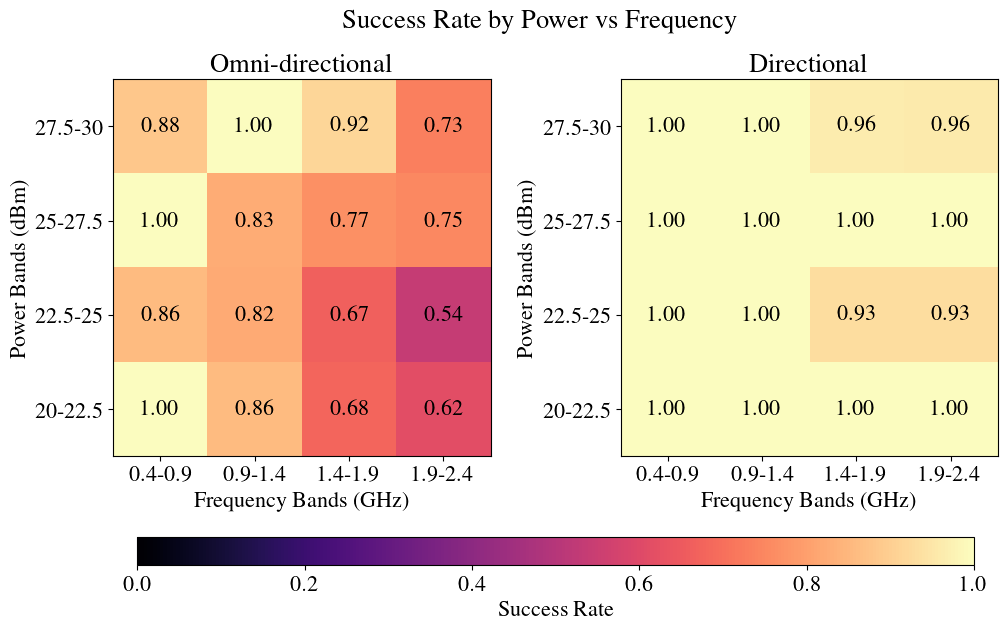}
    \caption{Localization success rate based on power and frequency of signal sources for each sensing strategy.}
    \label{fig:power_freq}
\end{figure}

Figure \ref{fig:power_freq} illustrates the effects of source power and frequency on emitter detection success rate across all trials for both sensing strategies (400 trials each). Fitting a logistic regression model to each dataset yields statistically significant ($p$-value $< 0.05$) effects for omnidirectional antennas. Source power shows a positive effect ($\beta_{x1} = +0.1207$), indicating that higher RSS values increase the likelihood of localization success. This result is intuitive: higher received power leads to more detections, enabling the formation of additional triangulation lines. Conversely, frequency shows a negative effect ($\beta_{x2} = -1.2744$), consistent with the expectation that higher frequencies suffer greater path loss, thereby reducing the probability of detection and the number of triangulation lines that can be generated. For directional sensing, neither power nor frequency exhibits a statistically significant effect on success rate. The higher antenna gain provides greater signal fidelity over longer ranges, making path loss far less influential at the tested spatial scale, regardless of frequency.

\begin{figure}
    \centering
    \includegraphics[width=1\linewidth]{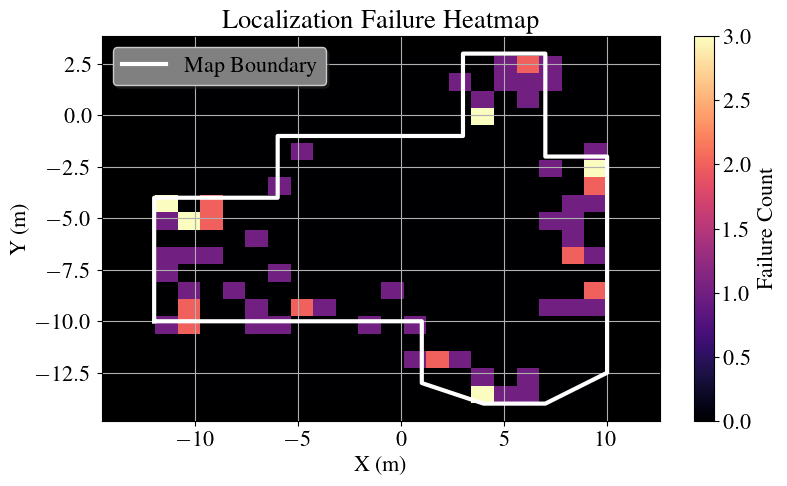}
    \caption{Heatmap of omnidirectional localization failures. Source locations are discretized into a 20$\times$20 grid. Failure counts of zero also include locations where no emitters were placed. Emitter locations are always inside the map.}
    \label{fig:fails}
\end{figure}

Examining omnidirectional points of failure more closely reveals that location is the dominant factor affecting localization success. Sources placed in remote or extreme areas of the map, such as corners, often go undetected, as shown in Figure \ref{fig:fails}. When combined with low-power, high-frequency transmitters, it becomes highly unlikely for these sources to be accurately located using the current method. Although concealed emitters in such locations may have limited coverage (e.g., audio monitoring devices), these areas are attractive hiding spots due to reduced foot traffic, therefore should not be overlooked.

\section{DISCUSSION}

Simulations demonstrate that geometric patrol route configurations for RSS-sensing mobile robots can be an effective strategy for localizing concealed emitters. In the approach described here, no prior knowledge of emitter parameters, such as location, power, frequency, transiency, path loss, or noise, are required. Our edge maxima triangulation method introduces a novel triangulation approach that analyses the spatial distribution of RSS measurements along patrol edges to identify local maxima, providing information about the approximate direction of the signal source. Differential evolution is used to generate patrol waypoints for a swarm of robots, optimizing map coverage given a defined sensing capability. Least-squares estimation of triangulation line intersections is then applied to predict the source location. Average localization errors range from 1.67--1.90 meters for omnidirectional sensing and 1.01--1.30 meters for directional sensing. 

An important result from this work is how the information gain of mobile RSS sensing robots is affected by the configuration of their patrol paths. Using geometric patterns enables robots to explicitly locate isotropic concealed emitters, independent of source parameters, with varying levels of success. For the same antenna and sensing range, triangular patrol routes are the most efficient per meter at generating regions of spatially explicit observation, compared to square and hexagonal formations. Furthermore, trial results shown in Figure \ref{fig:error} indicate that triangular routes produce lower and tighter localization error distributions for both omnidirectional and directional sensing. This is likely due to reduced ambiguity in the intersection of triangulation lines. As noted in \cite{Pack2009CooperativeTargets}, shallow orientation differences between triangulation lines negatively affect localization accuracy. Square and hexagonal patrol routes contain multiple parallel edges, leading to parallel triangulation lines and, consequently, less reliable intersection points. However, particularly for omnidirectional sensing, the success rate of triangular routes is significantly lower than that of other configurations. Although more efficient in terms of observation density, triangular patrols provide lower total area coverage for the same sensing range due to their edge length limitations. As a result, additional robots are required to sufficiently cover the entire map. There is a clear dependency of the robots' performance to localize concealed emitters on their interaction with the physical domain. It is here that the concept of spatial intelligence is shown to be an important, if not essential, factor to consider when deploying swarms of mobile electromagnetic sensing robots. 

\section{FUTURE WORK}

A divide-and-conquer strategy could improve accuracy by segmenting the map into manageable sub-regions. For instance, Voronoi partitioning divides the environment based on proximity to a set of predefined `sites'. Agent patrol routes can then be optimized for each resulting partition, and sub-regions covered sequentially until the entire map is patrolled. This method can be further adapted into an online, adaptive approach in which robots dynamically re-generate waypoints during the patrol cycle to cover potential blind spots left by previous routes. In addition to improving coverage, this introduces unpredictability, making it harder for transmitting adversarial devices to evade detection and thereby enhancing technical security. 

The results shown in Fig. \ref{fig:fails} motivate an investigation into incorporating secondary, heterogeneous behaviors within the swarm. Traditionally, swarms are composed of identical agents following the same local interaction rules and behaviors. However, studies such as \cite{York2024ShapingBehavior} highlight the benefits of ``functional heterogeneity" within a team of patrolling robots for anomaly detection, demonstrating how individual differences can improve collective performance. Applied to the methodology of this work, this could involve introducing specialized agents tasked solely with wall-following or corner-searching to increase the detectability of emitters in spatially restrictive areas. 

There is also scope to investigate the impact of inter-robot communication on localization predictions. In this study, a centralized aggregation of robot sensor data is assumed, enabling an accurate collective perception. In practice, however, communication ranges may be limited, and false positives (e.g., anomalous but harmless signals) may occur, creating an opportunity for consensus-based swarm monitoring (e.g., \cite{Madin2024}).

A notable limitation of this investigation is the absence of multi-path scattering. Multi-path effects are expected to significantly impact the applicability of the method in real-world implementations. Future work will incorporate advanced electromagnetic modeling capabilities (e.g., \cite{Pelham2023LyceanEM:Modelling, sionna}), within ROS2 physics-based simulation environments to accurately represent signal propagation and robot interactions. Advanced modeling will enable antenna array design, integration of spatial fingerprinting, and accurate channel modeling with beamforming \cite{PelhamSpatialAwareness}. 

Finally, closed-loop behavioral adaptation, as a method of spatial intelligence to investigate detected signals of interest, is a key capability to examine in future work.

\bibliographystyle{IEEEtran}
\bibliography{references}

\end{document}